\documentclass[review]{elsarticle}

\usepackage{lineno}
\modulolinenumbers[5]

\usepackage{amsmath,amssymb,amsfonts,bm}
\usepackage{caption}
\usepackage{multirow}
\usepackage{subfigure}
\usepackage{pifont}
\usepackage{float}
\usepackage{makecell}
\usepackage{booktabs}
\usepackage{multirow}
\usepackage{colortbl}
\usepackage[dvipsnames]{xcolor}
\usepackage{bbding}
\usepackage[ruled]{algorithm2e}

\usepackage[colorlinks, linkcolor=blue, anchorcolor=blue, citecolor=blue]{hyperref}
\usepackage{geometry}
\journal{Journal of Elsevier}

\begin{document}

\begin{frontmatter}

\title{Multi-objective Evolutionary Search of Variable-length Composite Semantic Perturbations}

\author[mysecondaryaddress]{Jialiang Sun}

\author[mysecondaryaddress]{Wen Yao\corref{mycorrespondingauthor}}
\ead{wendy0782@126.com}
\author[mysecondaryaddress]{Tingsong Jiang \corref{mycorrespondingauthor}}
\cortext[mycorrespondingauthor]{Corresponding author}
\ead{tingsong@pku.edu.cn}

%\author[mymainaddress]{Donghua Wang}
\author[mysecondaryaddress]{Xiaoqian Chen}

\address[mysecondaryaddress]{Defense Innovation Institute, Chinese Academy of Military Science, Beijing, 100071, China}
%\address[mymainaddress]{College of Computer Science and Technology, Zhejiang University}

\begin{abstract}
Deep neural networks have proven to be vulnerable to adversarial attacks in the form of adding specific perturbations on images to make wrong outputs. Designing stronger adversarial attack methods can help more reliably evaluate the robustness of DNN models. To release the harbor burden and improve the attack performance, auto machine learning (AutoML) has recently emerged as one successful technique to help automatically find the near-optimal adversarial attack strategy. However, existing works about AutoML for adversarial attacks only focus on $L_{\infty}$-norm-based perturbations. In fact, semantic perturbations attract increasing attention due to their naturalnesses and physical realizability. To bridge the gap between AutoML and semantic adversarial attacks, we propose a novel method called multi-objective evolutionary search of variable-length composite semantic perturbations (MES-VCSP). Specifically, we construct the mathematical model of variable-length composite semantic perturbations, which provides five gradient-based semantic attack methods. The same type of perturbation in an attack sequence is allowed to be performed multiple times. Besides, we introduce the multi-objective evolutionary search consisting of NSGA-II and neighborhood search to find near-optimal variable-length attack sequences. Experimental results on CIFAR10 and ImageNet datasets show that compared with existing methods, MES-VCSP can obtain adversarial examples with a higher attack success rate, more naturalness, and less time cost.
\end{abstract}

\end{frontmatter}

\section{Introduction}
Deep Neural Network (DNN) models have achieved great success in the majority of computer vision tasks \cite{szegedy2013deep,zhao2019object}. However, DNN proved to be vulnerable to specific perturbations, which can make models output wrong results \cite{goodfellow2014explaining,pgd2018}. The images with the perturbations are called adversarial examples (AEs). The phenomenon of AEs brings huge threats to real-world applications such as auto-driving and face recognition systems.

Focusing on the study of AEs, many works have been developed in the adversarial attack community. Developing stronger attack methods can approach the lower bound of the accuracy of DNN models, which can more reliably evaluate the model's robustness \cite{sun2023multi}. The goal of adversarial attacks is to generate specific perturbations to fool DNN models. According to the magnitude of the perturbations, existing adversarial attacks can be divided into $L_{p}$ norm-based like projection gradient descent (PGD) \cite{pgd2018,li2023bayesian,li2022approximated}
and unrestricted-based \cite{hosseini2018semantic,bhattad2019unrestricted,dreossi2018semantic,sun2023differential}. The former requires that the $L_{p}$ norm of perturbations that are added on original images (also called clean images) can not exceed a certain threshold. The latter does not limit the magnitude but needs to ensure that the semantics of clean images can not be changed, which is also called semantic perturbations. Due to the naturalness and physical realizability, semantic perturbations have been an emerging area, such as changing the geometry \cite{xiao2018spatially}, color \cite{shamsabadi2020colorfool}, and lightness \cite{sun2022ala} of clean images.

To enhance the robustness of deep neural network (DNN) models against adversarial examples, significant efforts have been devoted to improving the robustness of DNNs through various defense strategies. These strategies include adversarial training \cite{gowal2020uncovering}, defensive distillation \cite{papernot2016distillation}, dimensionality reduction \cite{bhagoji2017dimensionality}, input transformations \cite{guo2017countering}, and activation transformations \cite{dhillon2018stochastic}. The continuous development in the field of defense techniques has made it challenging to accurately evaluate the robustness of DNN models due to the limitations of existing adversarial attack algorithms. On the one hand, evaluating the robustness of models using adversarial attacks often requires high computational costs, making it computationally expensive to perform extensive evaluations. On the other hand, existing adversarial attack methods may not be able to fully explore the lower bound of the robust accuracy of defended models, thus providing an incomplete assessment of the model's robustness. Therefore, the evaluation of DNN model robustness remains a challenging task due to the trade-off between computational resources and the inability to obtain a comprehensive measure of a model's resilience against adversarial attacks. Further research and development are needed to overcome these challenges and establish more effective evaluation methodologies for assessing the robustness of DNN models.

To achieve a more reliable and efficient robustness evaluation of defended models, researchers have applied auto machine learning (AutoML) technique \cite{karmaker2021automl,he2021automl} into adversarial attack area for automated adversarial attacks. For instance, Francesco et al.  \cite{ReliableFrancesco} introduced auto attack (AA), an ensemble attack that combines four types of attacks to obtain lower robust accuracy against multiple defense methods. Tramer et al.  \cite{tramer2020adaptive} explored the impact of loss functions on adversarial attacks and demonstrated that suitable loss functions can further reduce the robust accuracy of defended models.
Mao et al. \cite{Composite2021} proposed a composite adversarial attack (CAA), which utilized the NSGA-II algorithm \cite{deb2002fast}  to search for near-optimal adversarial attacks while considering complexity and robust accuracy.
Furthermore, Yao \cite{Automated2021} proposed an adaptive auto attack, which not only searched for adversarial attack algorithms but also optimized hyperparameters, including randomization, using a hyperparameter optimization search method. Liu et al. \cite{liu2022practical} introduced an adaptive adversarial attack, which automatically selects the restart direction and attack budget for each image. In summary, the development of automated adversarial attack methods, such as auto attack, composite adversarial attack, adaptive auto attack, and adaptive adversarial attack, has demonstrated that conducting automated attacks for each defense model can yield lower robust accuracy and provide a more reliable evaluation of model robustness. These advancements contribute to a more comprehensive understanding of the effectiveness and limitations of defense strategies against adversarial attacks.

Though there exist some works about AutoML for $L_{\infty}$-norm-based attacks, it still remains an open problem about AutoML for unrestricted-based attacks. To improve the performance of semantic perturbations, Tsai et al. \cite{tsai2022towards,hsiung2022carben} proposed composite semantic perturbations (CSP), which combines the advantages of different semantic perturbations and form stronger attack algorithms. In their search space, five semantic perturbations are provided, including Hue, Saturation, Rotation, Contrast, and Brightness. Given one attack sequence, two modes can be selected, including the fixed and scheduled, where the fixed one denotes the fixed attack sequence, while the scheduled one would adjust the order of the attack sequence according to the batch data to perform stronger perturbations.

\begin{figure}[htbp]
	\centering
	\includegraphics[scale=0.3]{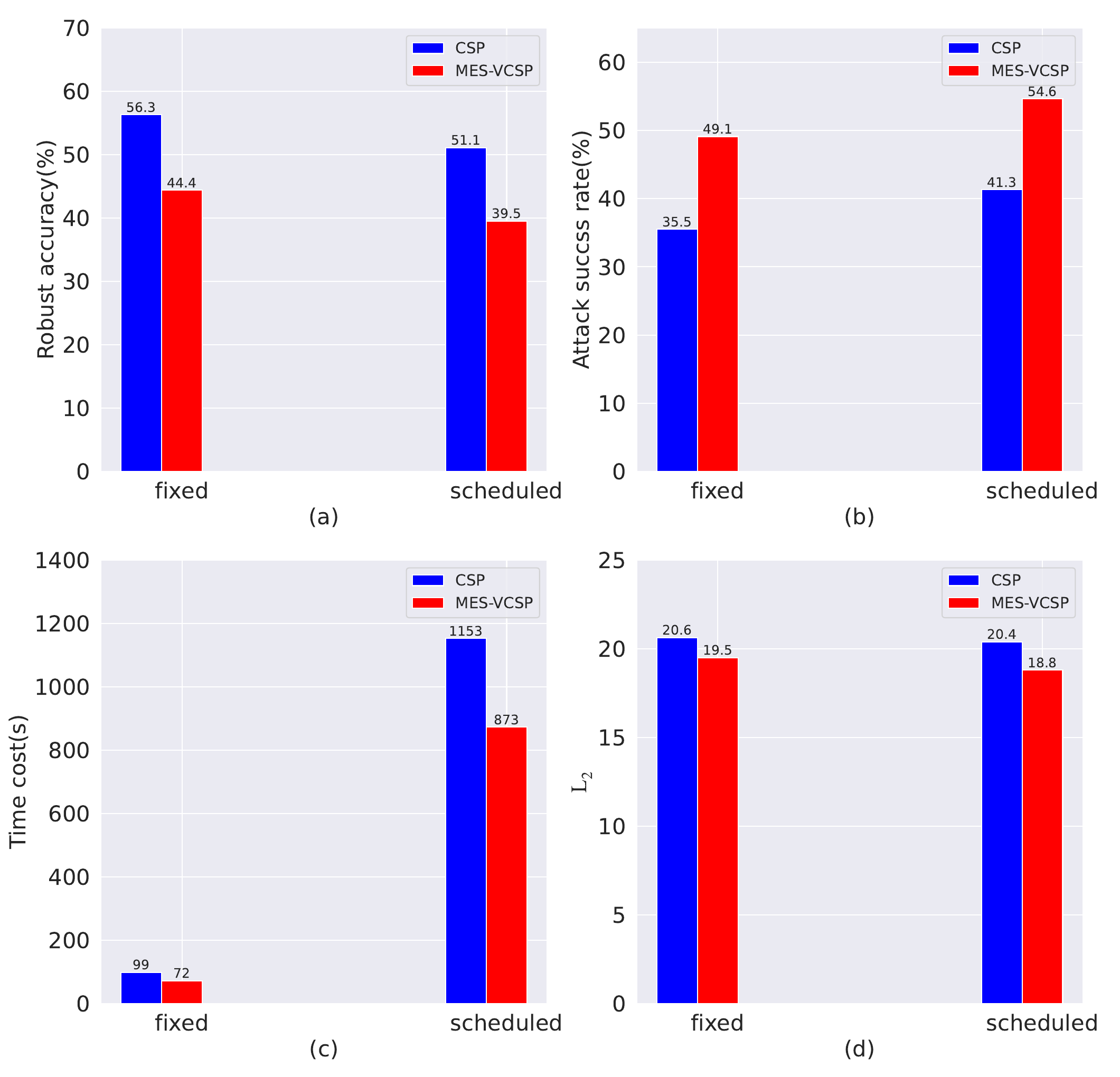}
	\caption{The comparison of the robust accuracy of the model, attack success rate, time cost, and $L_{2}$ distance between the generated AEs and clean images by CSP and MES-VCSP (ours) under two modes, including the fixed and scheduled.}
	
	\label{final}
\end{figure}

\begin{figure*}[h]
	\centering
	\includegraphics[scale=0.5]{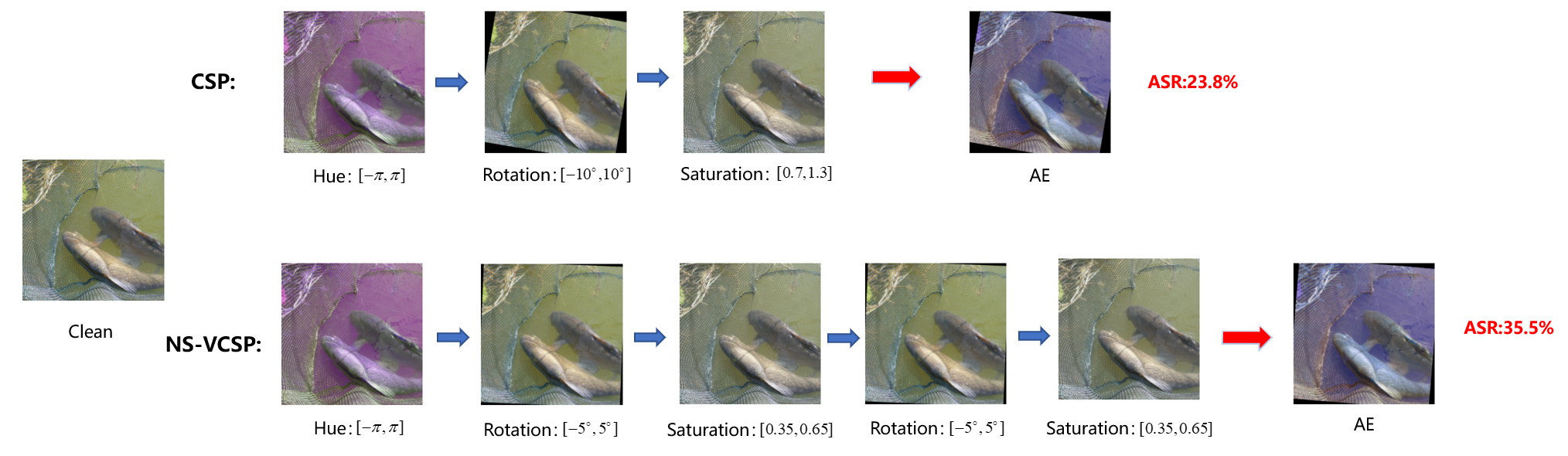}
	\caption{The illustration of the core idea and attack success rate (ASR) of our proposed NS-VCSP.}
	
	\label{core}
\end{figure*}

Based on their work on CSP, we find that the adversarial performance of CSP can be greatly improved by changing the length of attack sequences under the same search space. This phenomenon motivates us to propose the mathematical of variable-length composite semantic perturbations (VCSP), where the perturbation interval of each type of attack needs to be adaptively changed according to the number of the same attack in one attack sequence, making the final perturbations meet the original maximum constraints. In addition, to realize the automatical search, we introduce the multi-objective evolutionary search algorithm to find the near-optimal composite semantic perturbation sequence. Experimental results show that the attack performance can be improved greatly. An example of the comparison of CSP and the proposed MES-VCSP on the same adversarially trained model on the CIFAR10 dataset can be seen in Figure. \ref{final}. Our proposed MES-VCSP can achieve the attack success rate (ASR) of 49.1$\%$, while that of CSP is only 35.5$\%$. Besides, both the time cost and the $L_{2}$ distance between generated AEs and clean images of MES-VCSP are lower than CSP, showing the superiority of our proposed method. The core idea of VCSP can be seen in Figure. \ref{core}. By just changing the length of CSP from three to five, where both the rotation attack and saturation attack are performed twice, we can improve the ASR from 23.8$\%$ to 35.5$\%$ on the ImageNet dataset.

Our main contributions are concluded as follows:
\begin{itemize}
	\item We propose the mathematical model of variable-length composite semantic perturbations, where the interval of each type of perturbation is adaptively adjusted by the number in an attack sequence. 
	\item We introduce the multi-objective optimiztaion algorithm consisting of NSGA-II and neighborhood search algorithm to solve the proposed optimization problem, which can effectively find the near-optimal attack sequence.
	\item Compared with existing methods, experimental results on CIFAR10 and ImageNet datasets show that we can find the composite semantic perturbations with a higher attack success rate, less time cost, and more naturalness.
\end{itemize}

The remainder of this paper is organized as follows. Section \ref{sec2} introduces the background of composite semantic perturbations. Section \ref{sec3} elaborates our proposed method. Section \ref{exp} elaborates the experimental settings and results. 
Section \ref{conc} concludes this paper.
% The very first letter is a 2 line initial drop letter followed
% by the rest of the first word in caps.
% 
% form to use if the first word consists of a single letter:
% \IEEEPARstart{A}{demo} file is ....
% 
% form to use if you need the single drop letter followed by
% normal text (unknown if ever used by the IEEE):
% \IEEEPARstart{A}{}demo file is ....
% 
% Some journals put the first two words in caps:
% \IEEEPARstart{T}{his demo} file is ....
% 
% Here we have the typical use of a "T" for an initial drop letter
% and "HIS" in caps to complete the first word.

\section{Background}\label{sec2}

In this section, the background of five types of semantic perturbations is first introduced briefly. Then we describe the process of existing composite semantic perturbation. Finally, we elaborate on the concept of neighborhood search.

\subsection{Semantic Perturbations}
For most of the semantic perturbations, the
parameters that need to be optimized are continuous. Under the white-box attack setting, 
the parameters of semantic attacks can be updated by the gradient descent algorithm. Specifically, the paragraph outlines how the parameters can be updated for five different types of semantic perturbations, namely hue, saturation, brightness, contrast, and rotation. To optimize these perturbations, the iterative gradient sign method \cite{goyal2017accurate} is extended for $T$ iterations , which is defined as:

\begin{equation}
\delta_k^{t+1}=\operatorname{clip}_{\epsilon_k}\left(\delta_k^t+\alpha \cdot \operatorname{sign}\left(\nabla_{\delta_k^t} J\left(\mathcal{F}\left(A_k\left(X ; \delta_k^t\right)\right), y\right)\right)\right)
\label{sgd}
\end{equation}
where $t$ is the iteration number, $\mathcal{F}$ stands for the DNN model, $X$ denotes the input images, $y$ is the prediction label, $\epsilon_k$ is the perturbation interval. Assume $\epsilon_k = [\alpha_k, \beta_k]$, the element-wise clipping
operation $\operatorname{clip}_{\epsilon_k}$ is expressed as:

\begin{equation}
\operatorname{clip}_{\epsilon_k}(z)=\operatorname{clip}_{\left[\alpha_k, \beta_k\right]}(z)=\left\{\begin{aligned}
\alpha_k & \text { if } z<\alpha_k \\
z & \text { if } \alpha_k \leq z \leq \beta_k \\
\beta_k & \text { if } \beta_k<z
\end{aligned}\right.
\label{handle}
\end{equation}

The following description provides an explanation of each semantic attack.

\textbf{Hue}: The Hue attack can transfer clean images from RGB space to HSV space, causing a dropping in accuracy. The Hue value ranges from 0 to 2$\pi$. Hence, the maximum perturbation interval of Hue attack is [-2$\pi$,2$\pi$] \cite{tsai2022towards}.

\begin{equation}
x_H^t=\operatorname{Hue}\left(x_{\mathrm{adv}}^t\right)=\operatorname{clip}_{[0,2 \pi]}\left(x_H+\delta_H^t\right) 
\label{hue}
\end{equation}

\textbf{Saturation}: The Saturation attack can change the colorfulness of clean images by modifying the Saturation value, which ranges from 0 to 1. If the saturation value tends to be 1, the image becomes more colorful, while that is a gray-scale image if the saturation value is 0 \cite{tsai2022towards}.

\begin{equation}
x_S^t=\operatorname{Sat}\left(x_{\text {adv }}^t\right)=\operatorname{clip}_{[0,1]}\left(x_S \cdot \delta_S^t\right)  
\label{saturation}
\end{equation}

\textbf{Brightness and Contrast}: Brightness and contrast are different from hue and saturation in that they are defined in the RGB color space (pixel space) and determine the brightness and darkness differences of images. In our implementation, we first convert the images from the [0, 255] scale to the [0, 1] scale. The perturbation interval for brightness is defined as
$\epsilon_B = [\alpha_B, \beta_B]$, where $-1 \leq \alpha_B \leq \beta_B \leq 1$, while for contrast it is defined as   $\epsilon_C = [\alpha_C, \beta_C]$, where $-1 \leq \alpha_C \leq \beta_C \leq 1$.

Similarly to the hue and saturation attacks, we choose initial perturbations $\delta^{0}_{B}$ and $\delta^{0}_{C}$ uniformly from  $\epsilon_B$ and $\epsilon_C$, respectively, and update them using Eq. \ref{sgd}. The perturbed image $x^{t}_{adv}$ under the brightness attack is then obtained by adding the perturbation to the original image x, clamping the resulting values to be between 0 and 1, and scaling the result back to the [0, 255] scale.
The Brightness and Contrast attacks determine the lightness, darkness, and brightness difference of images in RGB color space \cite{tsai2022towards}, which can be obtained by Eq. \ref{brightness}.

\begin{equation}
x_{\mathrm{adv}}^t=\operatorname{clip}_{[0,1]}\left(x+\delta_B^t\right) \text { and } x_{\mathrm{adv}}^t=\operatorname{clip}_{[0,1]}\left(x \cdot \delta_C^t\right) 
\label{brightness}
\end{equation}

\textbf{Rotation}: The purpose of this transformation is to discover an angle of rotation that maximizes the loss of the rotated image. The rotation algorithm was developed by \cite{riba2020kornia}. Assuming we have a square image $x$, we can denote a pixel's position as $\left(i,j\right)$, and the center of $x$ as $\left(c,c\right)$. To calculate the new position $\left(i',j'\right)$ of a pixel rotated by an angle of $\theta$ degrees, we can use the following formula:

\begin{equation}
\left[\begin{array}{l}
i^{\prime} \\
j^{\prime}
\end{array}\right]=\left[\begin{array}{c}
\cos \theta \cdot i+\sin \theta \cdot j+(1-\cos \theta) \cdot c-\sin \theta \cdot c \\
-\sin \theta \cdot i+\cos \theta \cdot j+\sin \theta \cdot c+(1-\cos \theta) \cdot c
\end{array}\right]
\label{Rotation}
\end{equation}

\subsection{Composite Semantic Perturbations}

\begin{figure}[h]
	\centering
	\includegraphics[scale=0.65]{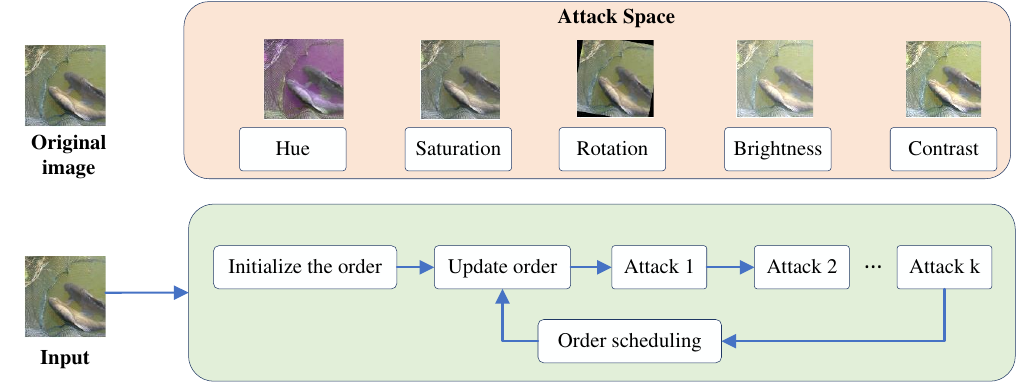}
	\caption{The illustration of composite semantic perturbation.}
	
	\label{csp_process}
\end{figure}

As for the white-box model, each attack above-mentioned can obtain the optimal perturbations by iteratively updating according to the gradient of model predictions. CSP tries to ensemble different types of semantic attacks to realize the more effective attacks. An illustration of the process of composite semantic perturbations is presented in Figure. \ref{csp_process}. The attack space includes five types of semantic adversarial attacks, namely Hue, Saturation, Rotation, Brightness, and Contrast. The configuration way of different types of attacks can be represented using the attack sequence, which consists of the types and order of multiple attacks.
When an attack sequence is given, the input image would be fed to different attacks subsequently. In the original work, CSP constructed the mathematical model of combining multiple types of attacks, which includes two modes, namely the fixed and scheduled.

\textbf{Fixed}: The fixed mode means that CSP utilizes the ensemble attack with the fixed sequence, where the order of attacks does not change during the process of attacking DNN models. 

\textbf{Scheduled}: The scheduled mode means that the order of the attack sequence can be adaptively adjusted with the batch data during the attacking process, which can achieve a higher ASR. The detailed way of updating the order of CSP can be seen in \cite{tsai2022towards}.

\subsection{Neighborhood Search}
To handle the discrete optimization problem, neighborhood
search emerges as an effective technique. Its core idea is to
construct the appropriate neighborhood for each candidate solution
according to the characteristic of the problem. The best
solutions to optimization problems can be obtained by iteratively
exploring the neighborhood. Over the past years, many
efforts have been devoted to developing effective neighborhood
search algorithms \cite{ahuja2002survey,pisinger2019large,mladenovic1997variable}.

\section{Method}\label{sec3}

\begin{figure*}[h]
	\centering
	\includegraphics[scale=0.85]{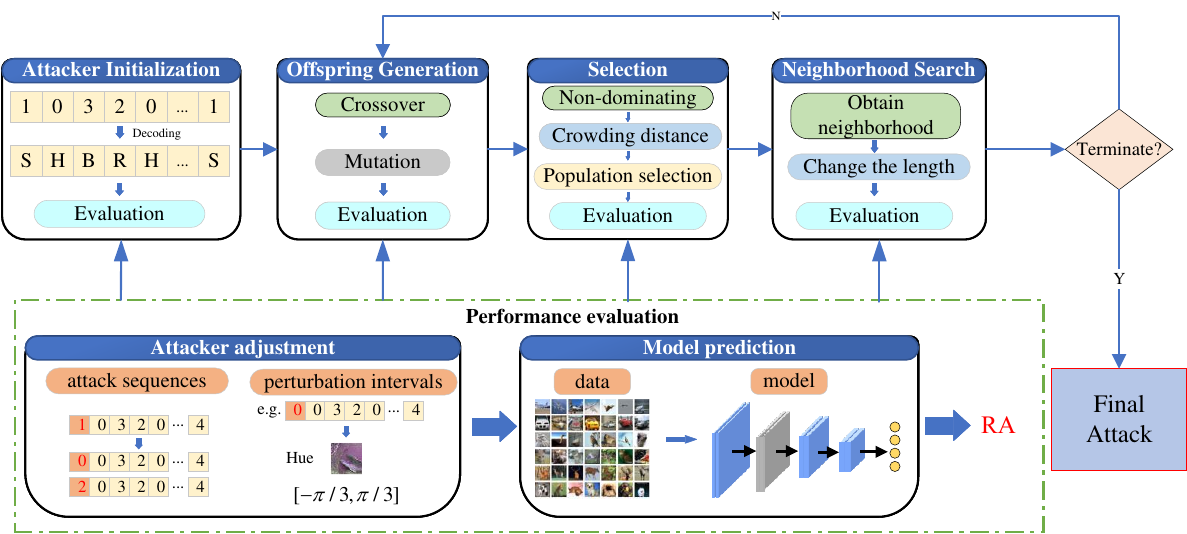}
	\caption{The process of multi-objective evolutionary search of variable-length composite semantic perturbations.}
	
	\label{process}
\end{figure*}

\begin{figure*}[h]
	\centering
	\includegraphics[scale=0.85]{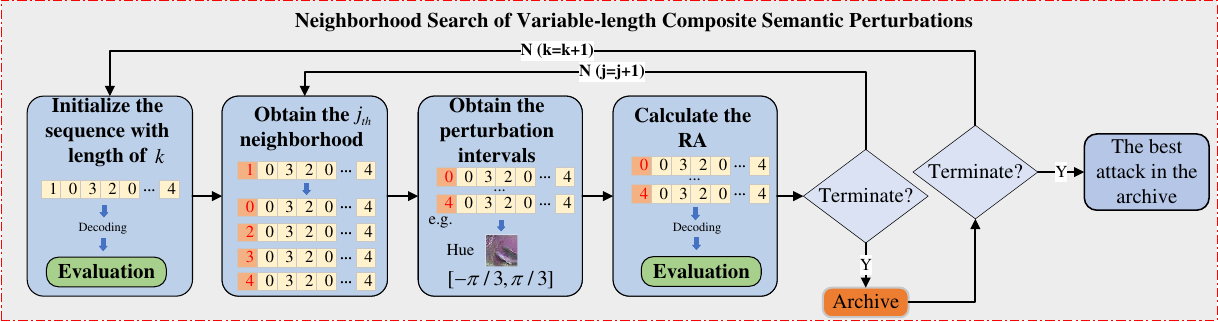}
	\caption{The process of neighborhood search.}
	
	\label{framework}
\end{figure*}

In this section, we introduce our proposed MES-VCSP method in detail. In section \ref{sec31}, we elaborate on the mathematical model of the proposed VCSP. In section \ref{sec32}, we introduce the multi-objective search algorithm for VCSP, including the search space, performance evaluation, and search strategy. The overview of our proposed MES-VCSP is presented in Figure. \ref{process}. Among that, the process of neighborhood search can be seen in Figure. \ref{framework}.

\subsection{The mathematical model of VCSP}\label{sec31}
In CSP, the length of attack sequence $k$ is fixed to the number of semantic perturbations in the search space $l_{sp}$, and each type of semantic perturbation is only allowed to be performed once. Then CSP tried to find the near-optimal configuration of different semantic perturbations under the scheduled mode. Different from CSP, we construct the mathematical model of variable-length composite semantic perturbations, which is presented as follows:
\begin{equation}
\begin{array}{ll}
\text {find} & \bm {x} \\
\min & F = [f_{1}(\bm {x}), f_{2}(\bm {x})] \\
\text { s.t. } & 1 \leq \bm {x}(i)) \leq l_{ sp }, \forall i\in [0,k]\\
& l_{ sp } \leq k \leq l_{ max } \\
\end{array}
\label{eq32}
\end{equation}
where $\bm x$ is the attack sequence of VCSP that we aim to find, $k$ denotes the length of VCSP, $l_{max}$ is the pre-settled maximum length of the attack sequence, $i$ represents the dimension of $\bm x$. From Eq. \ref{eq32}, we can see that our proposed VCSP can allow the same type of perturbation to be performed multiple times. To ensure the naturalness, the interval of each perturbation would be adaptively adjusted, where an example can be seen in \ref{321}. One optimization objective is the robust accuracy (RA) of the DNN model under adversarial attack $\bm x$, which is presented as follows.
\begin{equation}
f_{1} = Robust\text{ }Accuracy\text{ =}\frac{{{n}_{adv}}}{{{n}_{total}}}
\label{eqs}
\end{equation}
where ${n}_{total}$ stands for the number of total test samples that are generated by the adversarial attack. ${n}_{adv}$ denotes the number of examples that are predicted rightly. Another optimization objective $f_{2}$ is $L_{2}$ between the generated AEs and original images, which is calculated as follows.

\begin{equation}
\begin{array}{l}
%\mathcal{L}_{first} = \sum $MSE$(\mathcal{R}((\textbf{M},\textbf{T}_{g});\theta_{c}), I_{scene})
f_{2} = \mathcal{L}_{2} = \frac{\sum \limits_{i=1} {\text{MSE}(\mathbf{O}_{i},\mathbf{O}_{adv,i})}}{{n}_{total}}
\end{array}
\label{natural}
\end{equation}
where $\mathbf{O}_{i},\mathbf{O}_{adv,i}$ are $i_{th}$ original image and its corresponding adversarial example.

\subsection{Multi-objective search for VCSP}\label{sec32}
\subsubsection{search space}
\label{321}
To make a fair comparison, we still include five semantic perturbations provided by CSP in our search space. Denote the Hue, Saturation, Rotation, Brightness, and Contrast as 0, 1, 2, 3, 4, and we construct three search spaces, which are [0,1,2], [0,1,2,3], [0,1,2,3,4]. It means that the $n_{sp}$ in Eq. \ref{eq32} is 3, 4, and 5, respectively, in our defined search spaces. Under three search spaces, we search for the near-optimal configuration of semantic perturbations. The interval of each semantic perturbation is presented in Table \ref{interval_per}. In our variable-length CSP, the interval of each perturbation would be adaptively adjusted. Taking the Rotation operation as an example, if the attack sequence is [0,2,1,2], the interval of the Rotation operation at two attacks is [-5$^{\circ}$,5$^{\circ}$], which ensures that the final perturbations meet the original constraint [-10$^{\circ}$,10$^{\circ}$].

%\begin{table}[H]
%		\label{interval_per}
%	\begin{center}
%		\caption{Interval of each semantic perturbation} \label{tab:cap}
%		\begin{tabular}{|c|c|}
%			\hline
%			% after \\: \hline or \cline{col1-col2} \cline{col3-col4} ...
%			Peturbations & Interval
%			\\
%			\hline
%			Hue & [-$\pi$, $\pi$]  \\
%			Saturation& [0.7, 1.3] \\
%			Rotation & [-10, 10]  \\
%			Brightness& [-0.2, 0.2]\\
%			Contrast& [0.7, 1.3] \\
%			\hline
%		\end{tabular}
%	\end{center}
%\end{table}

\begin{table}[H]
\renewcommand\arraystretch{1.5}
\scriptsize
\centering
\caption{Interval of each semantic perturbation.}
\label{interval_per}
\setlength{\textwidth}{24mm}{
	\begin{tabular}{cccccc}
		
		\toprule
		
		\multicolumn{1}{c}{Peturbations} &Hue &Saturation&Rotation& Brightness&Contrast\\
		
		\midrule	
		Interval
		& [-$\pi$, $\pi$]  & [0.7, 1.3] & [-10$^{\circ}$, 10$^{\circ}$]  & [-0.2, 0.2]& [0.7, 1.3] \\
		
		\bottomrule		
\end{tabular}}
\end{table}

\begin{algorithm}[htbp]
\caption{Multi-objective Search of Variable-length Composite Semantic Perturbations}\label{algtotal}
\LinesNumbered
\KwIn{\\ The maximum number of iteration ${T}$, population size $n$, the training set $D_{train}$, the crossover rate $r_{c}$, the mutation rate $r_{m}$, the maximum length of the searched adversarial attack $l_{max}$;}
\KwOut{\\ The near-optimal adversarial attack.}
Generate the initial population $\bm{P}$  with $n$ composite semantic perturbations randomly. The length of each adversarial attack does not exceed $l_{max}$;
\\
Evaluate the RA and $L_{2}$ distance of each individual in  $\bm{P}$;\\

\While{$t \leftarrow 1:T$}{
	\textbf{// Evolution by discrete NSGA-II}\\
	Generate $n$ offspring individuals using crossover and mutation operation;\\
	Merge the offspring individuals with $\bm{P}$ and conduct the evaluation;\\
	Select $n$ individuals to form the new $\bm{P}$ using non-dominated-sorting and crowding distance;\\ 
	\textbf{// Neighborhood search}\\
	Obtain one individual in the first Pareto randomly $\bm{x}$;\\
	Conduct neighborhood search based on $\bm{x}$ according to \textbf{Algorithm} \ref{alg:3};\\
	Evaluate the individuals generated by local search and select the best one to substitute $\bm{x}$;\\	
}

Return The near-optimal adversarial attack
\end{algorithm}
\begin{algorithm}[htbp]
\caption{Generate the $j$th neighbour attack sequences solution $\bm x$:$ N(\bm x,j)=neighborhood( \bm x,j)$}
\label{alg:5}
\LinesNumbered
\KwIn{The attack sequence solution $\bm{x}$}
\KwOut{ The neighbour candidate solutions $ N(\bm x,j)$}
\For{$m \leftarrow 0: l_{sp}-1$}{
	$\bm x_{neighbor}$  $\leftarrow$ $\bm x$;\\
	\If {$x_{j} \neq m$} {
		$ x_{neighbor,j}$  $\leftarrow$ $m$;\\
		Include the new $\bm x_{neighbor}$ in $N( \bm x, j)$;\\
	}  
}

Return The neighbour candidate solutions $ N(\bm x,j)$
\end{algorithm}

\begin{algorithm}[htbp]
\caption{The pseud-code of neighborhood search}
\label{alg:3}
\LinesNumbered
\KwIn{The search space of semantic perturbations, maximum iteration number of outer loop $c$, an attack sequence $\bm x$, the maximum length of the searched adversarial attack $l_{max}$}
\KwOut{The optimal attack sequence $\bm x_{g}$}
$\bm x_{g} \leftarrow \bm x$;\\
$m \leftarrow$ Obtain current length of the attack sequence $\bm x_{g}$;\\
Calculate the RA and $L_{2}$ of $\bm x_{g}$;\\
Caulate the fitness of $\bm x_{g}$: $fitness_{g}$ = RA + $\lambda$$L_{2}$; \\
\For{$k \leftarrow m:min\left\{ m+c, l_{max} \right\}$}{
	
	\While{$flag = 1$}{
		$flag = 0$;\\
		\For{$j \leftarrow 1:k$}{
			
			Generate the neighbour solutions $N(\bm x_{g},j)$ according to \textbf{Algorithm} \ref{alg:5};\\
			Calculate the perturbation interval of each atttack sequence in $N(\bm x_{g},j)$; \\
			$f_{1}(N(\bm x_{g},j)), f_{2}(N(\bm x_{g},j))\leftarrow$ Calculate the RA and $L_{2}$ of $N(\bm x_{g},j)$ ;\\
			$f(\bm x) \leftarrow $ Calculate the fitness of $N(\bm x_{g},j)$;\\
			\If{ $max_{\bm x\in N{(\bm x_{g},j)}}$$f(\bm x)<fitness_{g}$}{
				$flag = 1$;\\
				$fitness_{g}$ $\leftarrow$ $min_{x\in N{(\bm x_{g},j)}}$;\\
				$\bm x_{g}$ $\leftarrow$ $argmin_{x\in N{(\bm x_{g},j)}}$;\\
			}
			
		}
	}
	Add the $\bm x_{g}$ to the set $\bm S$;\\
	$\bm x_{g} \leftarrow$ concat($\bm x_{g}$, random.randint(0, 5))
}
Return the best attack sequence $\bm x_{g}$ in set $\bm S$
\end{algorithm}
\subsubsection{performance evaluation}
During the search, we take the robust accuracy and $L_{2}$ distance as two metrics to evaluate the performance of each attack sequence. The robust accuracy is the percentage of images that can be correctly recognized by DNN models on total generated AEs. The $L_{2}$ distance denotes the $L_{2}$ norm between generated AEs and original clean images. Lower RA means better attack performance, while lower $L_{2}$ distance means better naturalness. In our proposed method, we aim to search for the near-optimal attack sequence with the lowest RA and smallest $L_{2}$ distance under the perturbation interval. If multiple attack sequences with different lengths possess the same RA, the short one would be selected as the final searched attack. It is because the short one would bring less computational cost. Our proposed search strategy includes NSGA-II and neighborhood search. The latter is a single-objective optimization algorithm, where the objective is formulated as follows.

\begin{equation}
\begin{array}{l}
%\mathcal{L}_{first} = \sum $MSE$(\mathcal{R}((\textbf{M},\textbf{T}_{g});\theta_{c}), I_{scene})
f = f_{1} + \lambda f_{2} = \text{RA} + \lambda L_{2}
\end{array}
\label{total}
\end{equation}

\subsubsection{search strategy}

To find the near-optimal attack sequence, we introduce the multi-objective search algorithm consisting of NSGA-II and neighborhood search. The whole process of our method is presented in  Algorithm \ref{algtotal}. We start by generating the initial population, including multiple random sequences, denoted as $\bm P$. Among the population, the length of composite semantic perturbations can differ. Thus a maximum length $n_{max}$ needs to be preset. After the population is evaluated on the dataset, the crossover and mutation operations are performed to generate the offspring. The crossover rate and mutation rate are set to $r_{c}$ and $r_{m}$, respectively. The new population is selected according to non-dominated-sorting. Then we will select the near-optimal individual randomly from the first Pareto to conduct the neighborhood search. Denote the selected individual as $\bm x$. The procedure of neighborhood search can be seen in Algorithm \ref{alg:3}. We first assign the value of $x$ to the $\bm x_{g}$. Then the following procedure consists of two loops. In the inner loop, we try to find the near-optimal sequence with the fixed length. We obtain the neighborhood by changing one value at each time according to the dimensionality, as presented in Algorithm \ref{alg:5}.  All the attack sequences in the neighborhood would be evaluated and compared with $\bm x_{g}$. The best one would be preserved as new $\bm x_{g}$. If all the dimensionalities are changed and do not improve the attack performance, the algorithm will conduct the outer loop. In the outer loop, we try to find the near-optimal length of attack sequences from short to long. The initial length is the original length of $x$, which is denoted as $m$. The maximum length is set to $min\left\{ m+c, n_{max} \right\}$, where $c$ is the maximum iteration number of the outer loop and $l_{max}$ is the maximum length of the searched adversarial attack. By $min\left\{ m+c, n_{max} \right\}$, we can not only set the maximum length of the attack sequence but also limit the computational cost by setting $c$ flexibly. After each inner loop terminates, the best attack sequence would be preserved to an archive $\bm S$ until the outer loop ends. Then the best attack sequence in $\bm S$ would be the final searched attack. By neighborhood search, the promising solution searched by NSGA-II can efficiently find the better attack sequence. Thus, the whole algorithm can achieve a better balance between the exploitation ability and exploration ability.

\section{Experimental Results}
\label{exp}

\subsection{Experimental settings}

In our experimental settings, we select the CIFAR10 and ImageNet datasets. The number of test images is set to 10,000 and 500, respectively. The threat models include standard training models and defensed models. The standard training models are VGG19, ResNet18, GoogleNet, DenseNet121, and MobileNetV2. On both datasets, the attack is searched on the former model (source model) and transferred to the latter one (target model), which is to illustrate the transferability of the searched attack and avoid the repeatedly search. In our experiments, VGG19 is set as the source model. In addition, we also transfer the searched attacks on CIFAR10 and ImageNet datasets to the defensed models to evaluate the performance of the proposed method. For the defensed models, we utilize generalized adversarial training (GAT), ensemble adversarial training (EnsembleAT), label smoothing (LS), fast adversarial training (FastAT), and robust transfer (RT). For the CIFAR10 dataset, we retrain the VGG19, ResNet18, GoogleNet, DenseNet121, and MobileNetV2 models using the GAT technique. For the ImageNet dataset, five model architectures, including WideResnet50, WideResnet50, Resnet50,  Inception$\_$ResNet$\_$V2, and  Inception$\_$V3, adopt the corresponding above-mentioned defensed strategy.

\textbf{Implementation details}: 
To provide a comprehensive comparison, we include the same search space, [0,1,2,3,4], as CSP. In the MES-VCSP, we just set the iteration number of each attack component as 1. The maximum length of VCSP $l_{max}$ is set to 8. To provide a fair comparison, in CSP, the corresponding iteration number of each attack component is set to 2. During the search for the near-optimal attack sequence, we take the robust accuracy and the $L_{2}$ distance between the generated AEs and clean images as the objectives. Both the population size $n$ and maximum number of iteration $T$ are set to 20. The crossover rate and mutation rate, namely $r_{c}$ and $r_{m}$, are set to 0.6 and 1, respectively. Duing the neighborhood search, the coefficient $\lambda$ in Eq. \ref{total} is set to 1. The maximum iteration number of outer loop $c$ is set to 1. The maximum length of the searched adversarial attack $l_{max}$ is set to 8. The number of images in the training dataset to evaluate the performance of attack sequences for CIFAR10 and ImageNet are set to 500 and 100, respectively. When the searched attack is evaluated, we also include other metrics, such as the attack success rate and time cost of the evaluation. All the experiments are conducted using a single NVIDIA Titan P2 GPU.

\begin{table*}[htbp]
\renewcommand\arraystretch{1.5}
\scriptsize
\centering
\caption{The comprehensive comparison of CSP and MES-VCSP on the CIFAR10 dataset.}
\label{search_time}
\resizebox{\textwidth}{!}{
	\begin{tabular}{c|c|ccc|ccccc}
		
		\toprule
		
		\multirow{1}{*}{Model}&\multirow{1}{*}{Clean }&\multirow{1}{*}{Method}&	\multirow{1}{*}{Attack sequence}&\multirow{1}{*}{Mode}&	\multirow{1}{*}{Robust Accuracy ($\downarrow$)}&\multicolumn{1}{c}{ASR ($\uparrow$)} &\multicolumn{1}{c}{Time cost ($\downarrow$)} &\multicolumn{1}{c}{$L_{2}$ ($\downarrow$)}\\
		
		\midrule			
		&&CSP	&[0,1,2,3,4]& fixed &    43.3$\%$&53.4$\%$&50.61s&12.70 \\	
		VGG19&92.34	 &MES-VCSP&[2,1,0,1,4,2,3]	& fixed  & \textbf{30.1}$\%$&\textbf{67.7}$\%$&\textbf{45.64}s&\textbf{12.61} \\
		&&CSP&	[0,1,2,3,4]&  scheduled &  20.6$\%$&77.7$\%$&380.63s&14.32 \\	
		&	&MES-VCSP		  & 	[2,1,0,1,4,2,3]  &scheduled &\textbf{13.7}$\%$&\textbf{85.2}$\%$&\textbf{315.29}s&\textbf{13.38} \\
		
		\midrule			
		&&CSP	&[0,1,2,3,4]& fixed &  51.6$\%$&45.6$\%$&60.73s&13.13   \\	
		ResNet18&94.87	 &MES-VCSP&[2,1,0,1,4,2,3]	 & fixed  & \textbf{33.6}$\%$&\textbf{64.6}$\%$&\textbf{49.61}s&\textbf{13.12}\\
		&&CSP&	[0,1,2,3,4]&  scheduled &   27.6$\%$&70.9$\%$&544.46s&14.94\\	
		&	&MES-VCSP		  & 	[2,1,0,1,4,2,3] &scheduled &\textbf{15.7}$\%$&\textbf{83.5}$\%$&\textbf{438.91}s&\textbf{13.95}\\
		
		\midrule			
		&&CSP	&[0,1,2,3,4]& fixed &    53.0$\%$&44.5$\%$&183.93s&13.38\\	
		GoogleNet&95.26	 &MES-VCSP&	[2,1,0,1,4,2,3] & fixed  & \textbf{32.3}$\%$&\textbf{66.3}$\%$&\textbf{142.76}s&\textbf{13.00} \\
		&&CSP&	[0,1,2,3,4]&  scheduled &  27.7$\%$&70.9$\%$&2043.92s&15.18 \\	
		&	&MES-VCSP		  &[2,1,0,1,4,2,3]   &scheduled &\textbf{15.2}$\%$&\textbf{84.1}$\%$&\textbf{1568.07}s&\textbf{14.04}\\
		
		\midrule			
		&&CSP	&[0,1,2,3,4]& fixed &    49.1$\%$&48.0$\%$&202.31s&13.45\\	
		DenseNet121&94.43	 &MES-VCSP& [2,1,0,1,4,2,3] &fixed & \textbf{32.2}$\%$&\textbf{66.1}$\%$&\textbf{154.31}s&\textbf{13.09}\\
		&&CSP&	[0,1,2,3,4]&  scheduled &  25.0$\%$&73.5$\%$&2164.73s&15.16 \\	
		&	&MES-VCSP		  & [2,1,0,1,4,2,3]	 &scheduled &\textbf{15.4}$\%$&\textbf{83.7}$\%$&\textbf{1663.82}s&\textbf{13.96}\\
		
		\midrule			
		&&CSP	&[0,1,2,3,4]& fixed & 43.9$\%$ & 53.0$\%$&86.23s&12.84   \\	
		MobileNetV2&93.01	 &MES-VCSP&	[2,1,0,1,4,2,3]& fixed  & \textbf{27.3}$\%$&\textbf{70.9}$\%$&\textbf{66.12}s&\textbf{12.31}  \\
		&&CSP&	[0,1,2,3,4]&  scheduled &  19.2$\%$&79.4$\%$&815.75s&14.33 \\	
		&	&MES-VCSP		  & 	[2,1,0,1,4,2,3] &scheduled & \textbf{10.5}$\%$&\textbf{88.7}$\%$&\textbf{633.81}s&\textbf{13.22}\\

		\bottomrule		
\end{tabular}}
\end{table*}

\begin{table*}[htbp]
\renewcommand\arraystretch{1.5}
\scriptsize
\centering
\caption{The comprehensive comparison of CSP and MES-VCSP on the ImageNet dataset.}
\label{Imagenet}
\resizebox{\textwidth}{!}{
	\begin{tabular}{c|c|ccc|ccccc}
		
		\toprule
		
		\multirow{1}{*}{Model}&\multirow{1}{*}{Clean }&\multirow{1}{*}{Method}&	\multirow{1}{*}{Attack sequence}&\multirow{1}{*}{Mode}&	\multirow{1}{*}{Robust Accuracy ($\downarrow$)}&\multicolumn{1}{c}{ASR ($\uparrow$)} &\multicolumn{1}{c}{Time cost ($\downarrow$)} &\multicolumn{1}{c}{$L_{2}$ ($\downarrow$)}\\
		
		\midrule			
		&&CSP	&[0,1,2,3,4]& fixed &     6.8$\%$&88.4$\%$&23.73s&86.15\\	
		VGG19&58.4$\%$	 &MES-VCSP&[3,4,2,0,1,3,2,1]	& fixed  & \textbf{1.6}$\%$&\textbf{97.2}$\%$&\textbf{20.51}s&\textbf{55.56}\\
		&&CSP&	[0,1,2,3,4]&  scheduled & 1.6$\%$&97.2$\%$&75.51s&71.19\\	
		&	&MES-VCSP		  & 	[3,4,2,0,1,3,2,1]  &scheduled & \textbf{1.3}$\%$&\textbf{97.8}$\%$&\textbf{49.84}s&\textbf{62.58}\\
		
		\midrule			
		&&CSP	&[0,1,2,3,4]& fixed &  3.9$\%$&93.5$\%$&9.44s&84.00   \\	
		ResNet18&59.4$\%$	 &MES-VCSP&[3,4,2,0,1,3,2,1]	 & fixed  & \textbf{1.9}$\%$&\textbf{96.7}$\%$&\textbf{9.12}s&\textbf{61.32}\\
		&&CSP&	[0,1,2,3,4]&  scheduled &   0.3$\%$&99.5$\%$&20.03s&82.39\\	
		&	&MES-VCSP		  & 	[3,4,2,0,1,3,2,1] &scheduled &\textbf{0.0}$\%$&\textbf{100}$\%$&\textbf{14.85}s&\textbf{65.01}\\
		
		\midrule			
		&&CSP	&[0,1,2,3,4]& fixed &   21.6$\%$&70.9$\%$&16.53s&94.97 \\	
		GoogleNet&74.2$\%$ &MES-VCSP&	[3,4,2,0,1,3,2,1] & fixed  & \textbf{9.0}$\%$&\textbf{87.8}$\%$&\textbf{14.56}s&\textbf{77.37} \\
		&&CSP&	[0,1,2,3,4]&  scheduled &  4.8$\%$&93.5$\%$&90.26s&84.72\\	
		&	&MES-VCSP		  &[3,4,2,0,1,3,2,1]   &scheduled &\textbf{1.6}$\%$&\textbf{97.8}$\%$&\textbf{50.75}s&\textbf{73.59}\\
		
		\midrule			
		&&CSP	&[0,1,2,3,4]& fixed &  12.3$\%$&84.1$\%$&31.05s&90.47  \\	
		DenseNet121&75.2$\%$	 &MES-VCSP& [3,4,2,0,1,3,2,1] &fixed & \textbf{3.5}$\%$&\textbf{95.3}$\%$&\textbf{23.78}s&\textbf{80.57}\\
		&&CSP&	[0,1,2,3,4]&  scheduled &  2.9$\%$&96.1$\%$&127.06s&\textbf{85.85}\\	
		&	&MES-VCSP		  & [3,4,2,0,1,3,2,1]	 &scheduled &\textbf{0.6}$\%$&\textbf{99.1}$\%$&\textbf{59.70}s&87.05\\
		
		\midrule			
		&&CSP	&[0,1,2,3,4]& fixed & 5.5$\%$&90.5$\%$&12.28s&84.14  \\	
		MobileNetV2&54.2$\%$	 &MES-VCSP&	[3,4,2,0,1,3,2,1]& fixed  &  \textbf{2.3}$\%$&\textbf{95.8}$\%$&\textbf{10.76}s&\textbf{53.79} \\
		&&CSP&	[0,1,2,3,4]&  scheduled & \textbf{0.6}$\%$&\textbf{98.8}$\%$&23.59s&69.14 \\	
		&	&MES-VCSP		  & 	[3,4,2,0,1,3,2,1] &scheduled & 1.0$\%$&98.2$\%$&\textbf{19.19}s&\textbf{52.46}\\

		\bottomrule		
\end{tabular}}
\end{table*}

\subsection{The performance MES-VCSP on standard training models}
\begin{figure}[htbp]
\centering
\includegraphics[scale=0.7]{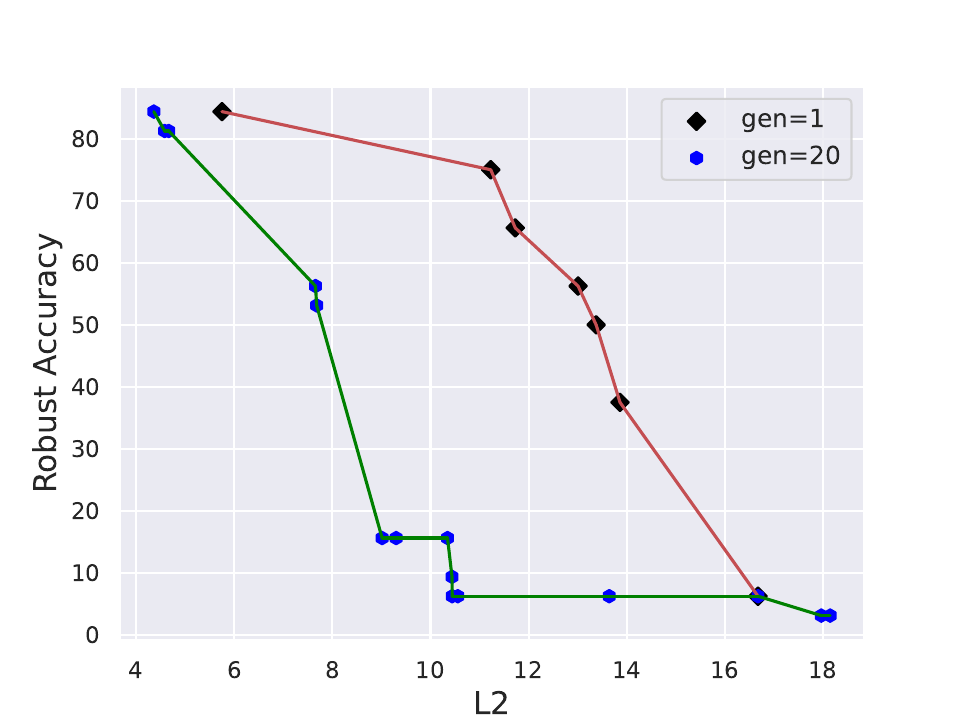}

\caption{The visualization of population after 20 iterations.}

\label{results}
\end{figure}

\begin{figure}[h]
\centering
\includegraphics[scale=0.25]{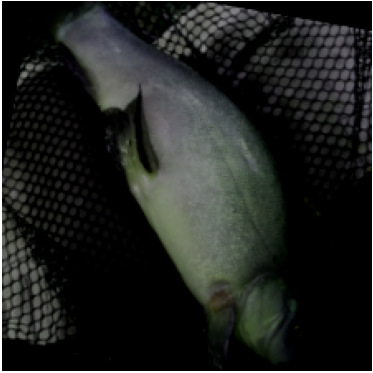}
\includegraphics[scale=0.25]{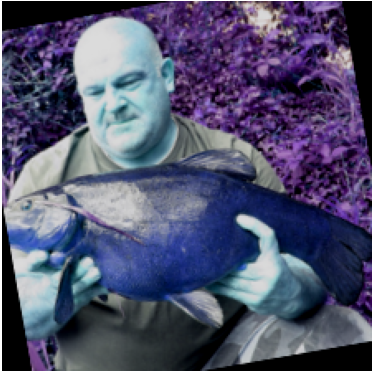}
\includegraphics[scale=0.25]{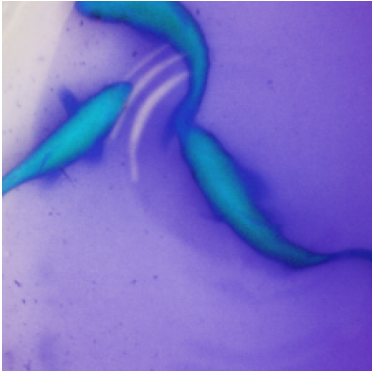}
\includegraphics[scale=0.25]{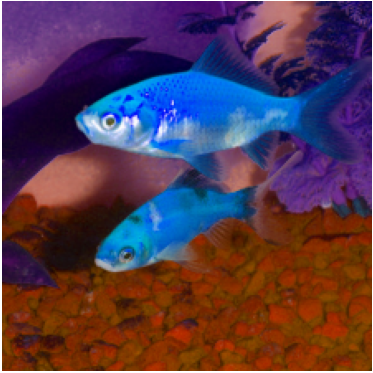}
\includegraphics[scale=0.25]{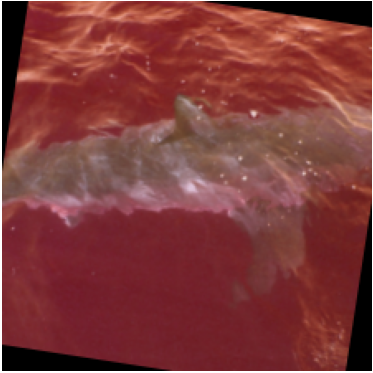}

\includegraphics[scale=0.25]{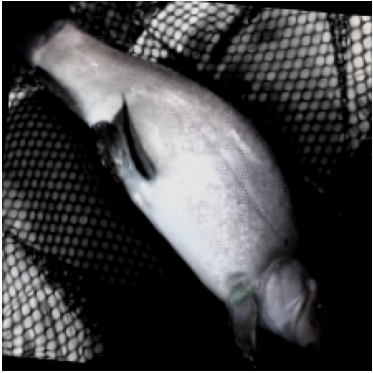}
\includegraphics[scale=0.25]{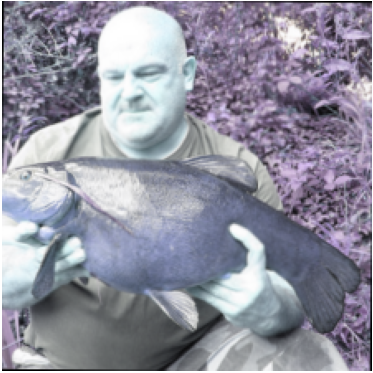}
\includegraphics[scale=0.25]{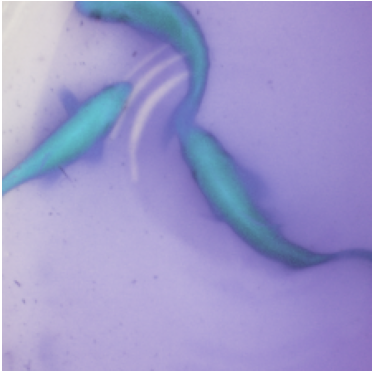}
\includegraphics[scale=0.25]{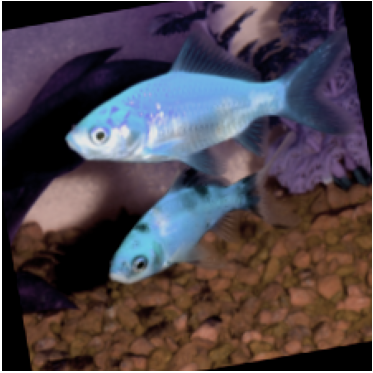}
\includegraphics[scale=0.25]{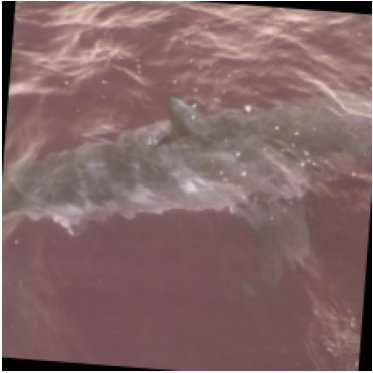}
\caption{The visualization of generated AEs by CSP (first row) and  MES-VCSP (second row).}

\label{VISUAL}
\end{figure}

We compare the performance of CSP and MES-VCSP on two modes, including the fixed and scheduled. The search Pareto front using VGG19 as the source model is presented in Figure. \ref{results}. From Figure. \ref{results}, we can see that the proposed optimization model is a typical multi-objective optimization problem. All the individuals in the Pareto front are the optimal solutions. We compromise by choosing a solution with relatively good robust accuracy and $L_{2}$ distance as the searched attack to evaluate other models. The searched attacks on the CIFAR10 dataset and ImageNet dataset are [2,1,0,1,4,2,3] and [3,4,2,0,1,3,2,1], respectively. The evaluation results of five standard training models under CSP and the searched attacks are shown in Table \ref{search_time} and Table \ref{Imagenet}, respectively. From Table \ref{search_time} and Table \ref{Imagenet}, we can see that in almost all cases, MES-VCSP can obtain a significantly higher ASR than CSP. In some cases, like GoogleNet in CIFAR10 dataset, the improvement of ASR can even reach over 20$\%$. Besides, in all cases, the time cost of MES-VCSP is less than CSP, which verifies the effectiveness of our proposed method. The $L_{2}$ distances of generated AEs of MES-VCSP are also smaller than CSP in most cases. Some examples of CSP and MES-VCSP are presented in Figure. \ref{VISUAL}. From Figure. \ref{VISUAL}, we can see that the AEs generated by MES-VCSP can still possess good natural appearances and original semantics. It shows that our improvement in ASR does not sacrifice the semantic information of the original images. In addition, experimental results also show that the searched attack possesses superior transferability across different models. In all cases of target models, the ASR of MES-VCSP is also higher than CSP. It illustrates that we do not need to search for the near-optimal attack sequence repeatedly from scratch for each model, further showing the effectiveness of our proposed method. In general, our proposed method can obtain the AEs with higher ASR, less time cost, and more naturalness.

%Some of the reasons that our proposed method is so effective is that by dividing one type of semantic perturbation into attack components multiple times, we explore the larger design spaces of the optimization problem than CSP.

%\begin{figure*}[t]
%	\centering
%	\includegraphics[scale=0.5]{gen_0_caa.jpg}	
%	\includegraphics[scale=0.5]{gen_19_caa.jpg}
%	
%	\caption{The visualization of population during optimization (the left figure is the population initialization, and the right one is the population after 20 iterations).}
%	
%	\label{results}
%\end{figure*}

%\subsection{The performance MES-VCSP on ImageNet dataset}

\subsection{The performance of MES-VCSP on defensed models}
\begin{figure*}[htbp]
\centering
\includegraphics[scale=0.3]{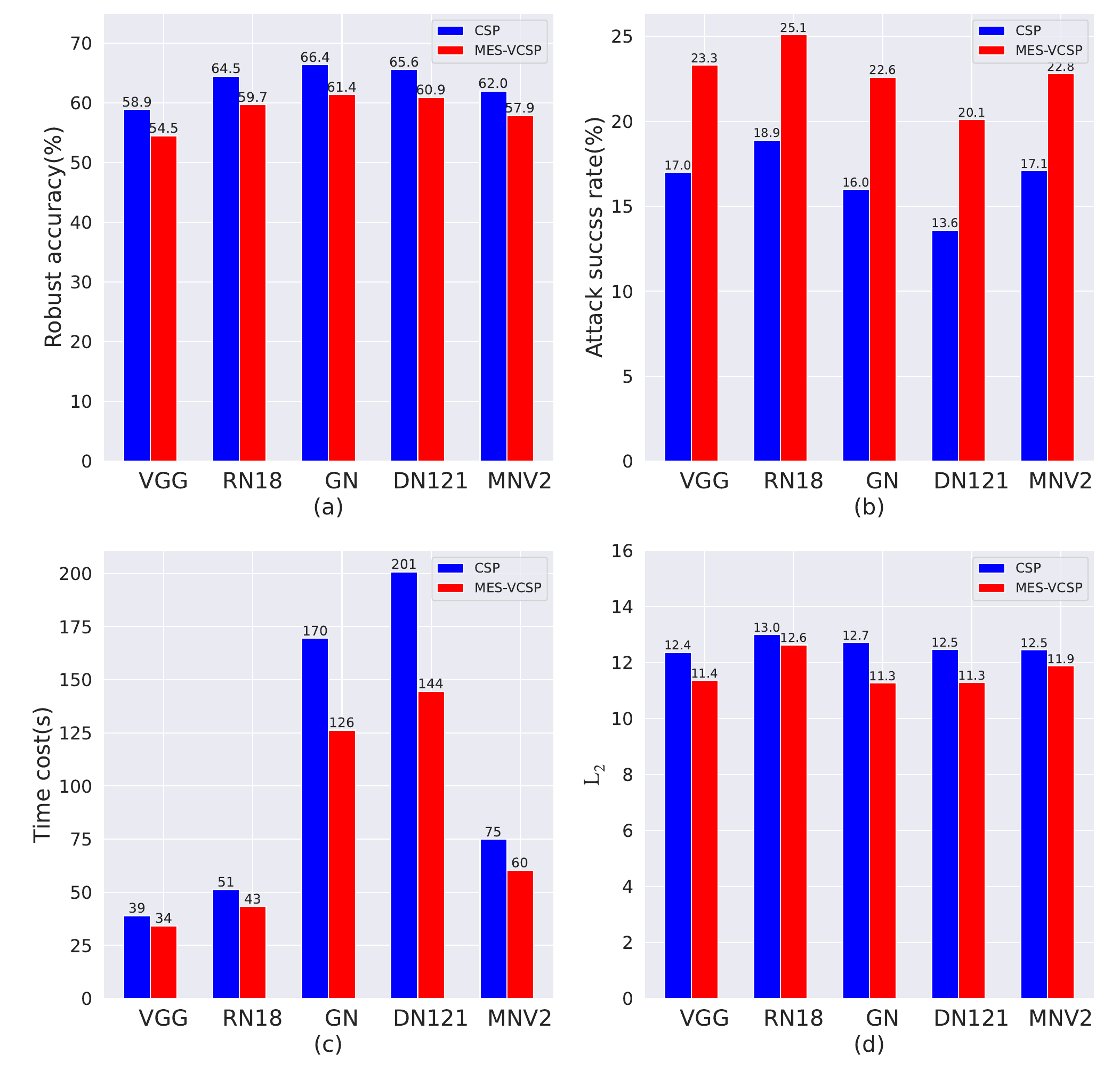}
\caption{The comparison of CSP and MES-VCSP on defensed model on the CIFAR10 dataset under the fixed mode. VGG, RN18, GN, DN121, and MNV2 denote VGG19, ResNet18, GoogleNet, DenseNet121, and MobileNetV2.}

\label{defensecifar10}
\end{figure*}
\begin{figure*}[htbp]
\centering
\includegraphics[scale=0.3]{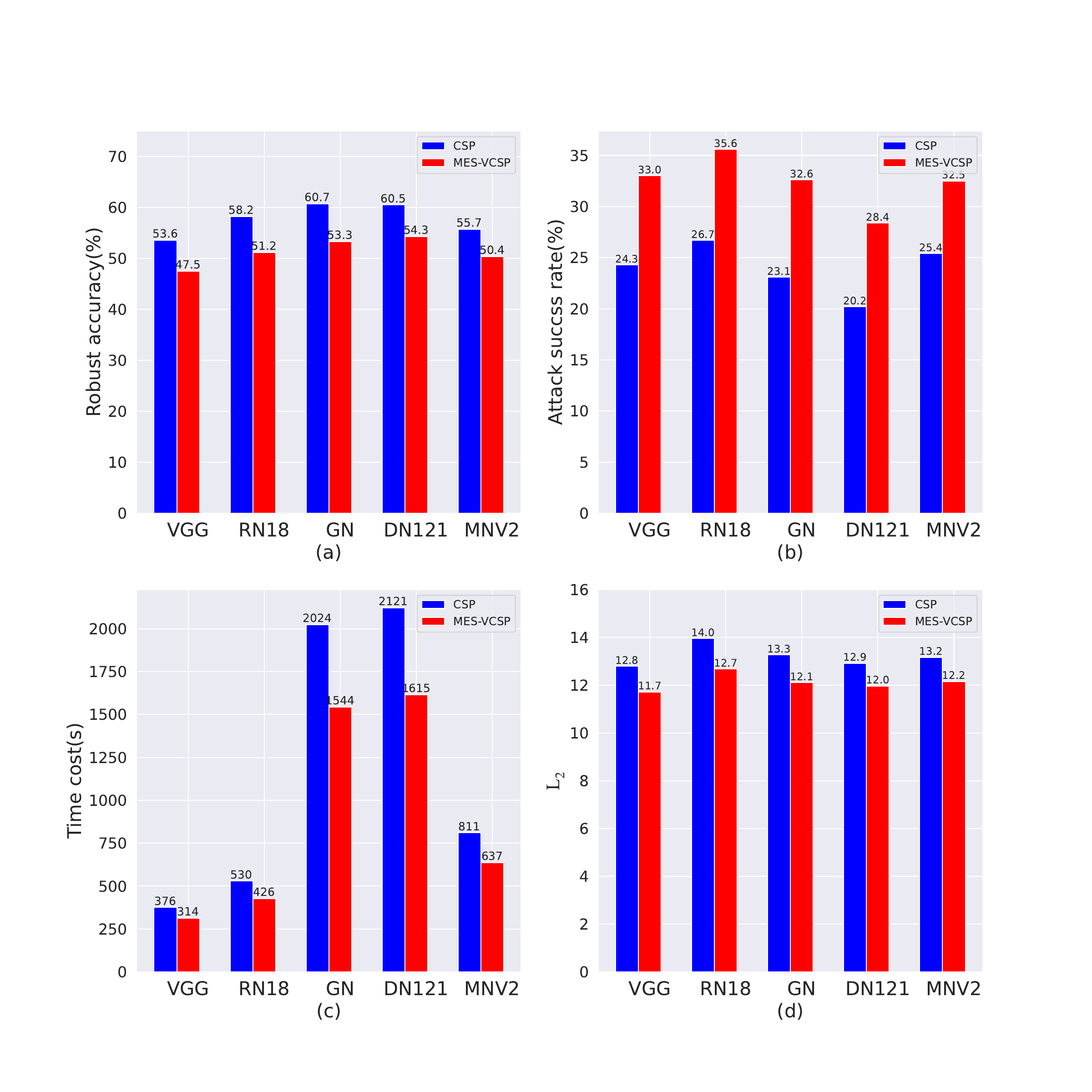}
\caption{The comparison of CSP and MES-VCSP on defensed model on the CIFAR10 dataset under the scheduled mode.}

\label{defensecifar10sche}
\end{figure*}

\begin{figure*}[htbp]
\centering
\includegraphics[scale=0.3]{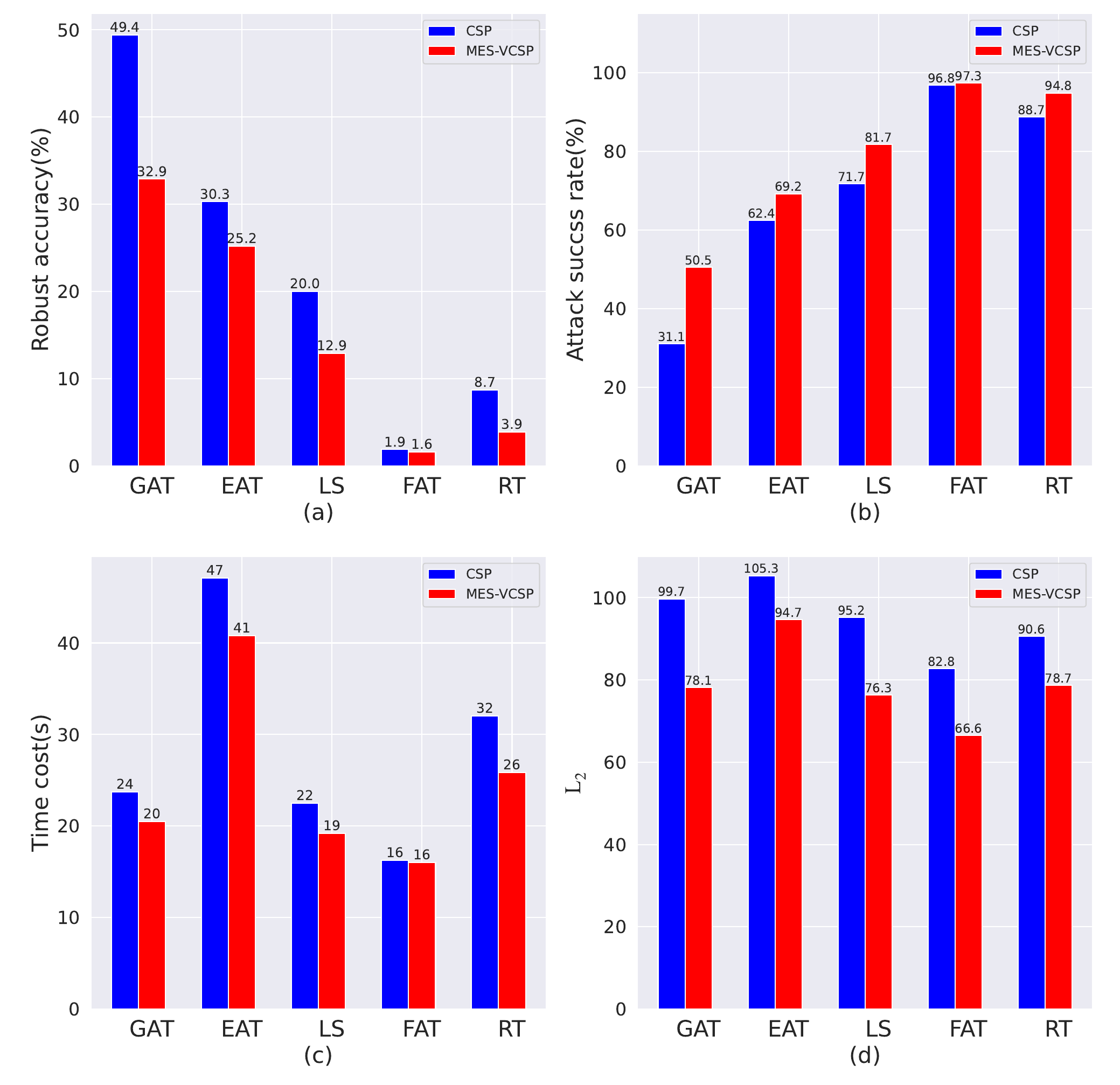}
\caption{The comparison of CSP and MES-VCSP on defensed model on the ImageNet dataset under the fixed mode. GAT, EAT, LS, FAT, and RT denote generalized adversarial training, ensemble adversarial training, label smoothing, fast adversarial training, and robust transfer.}

\label{defenseimagenet}
\end{figure*}

\begin{figure*}[htbp]
\centering
\includegraphics[scale=0.3]{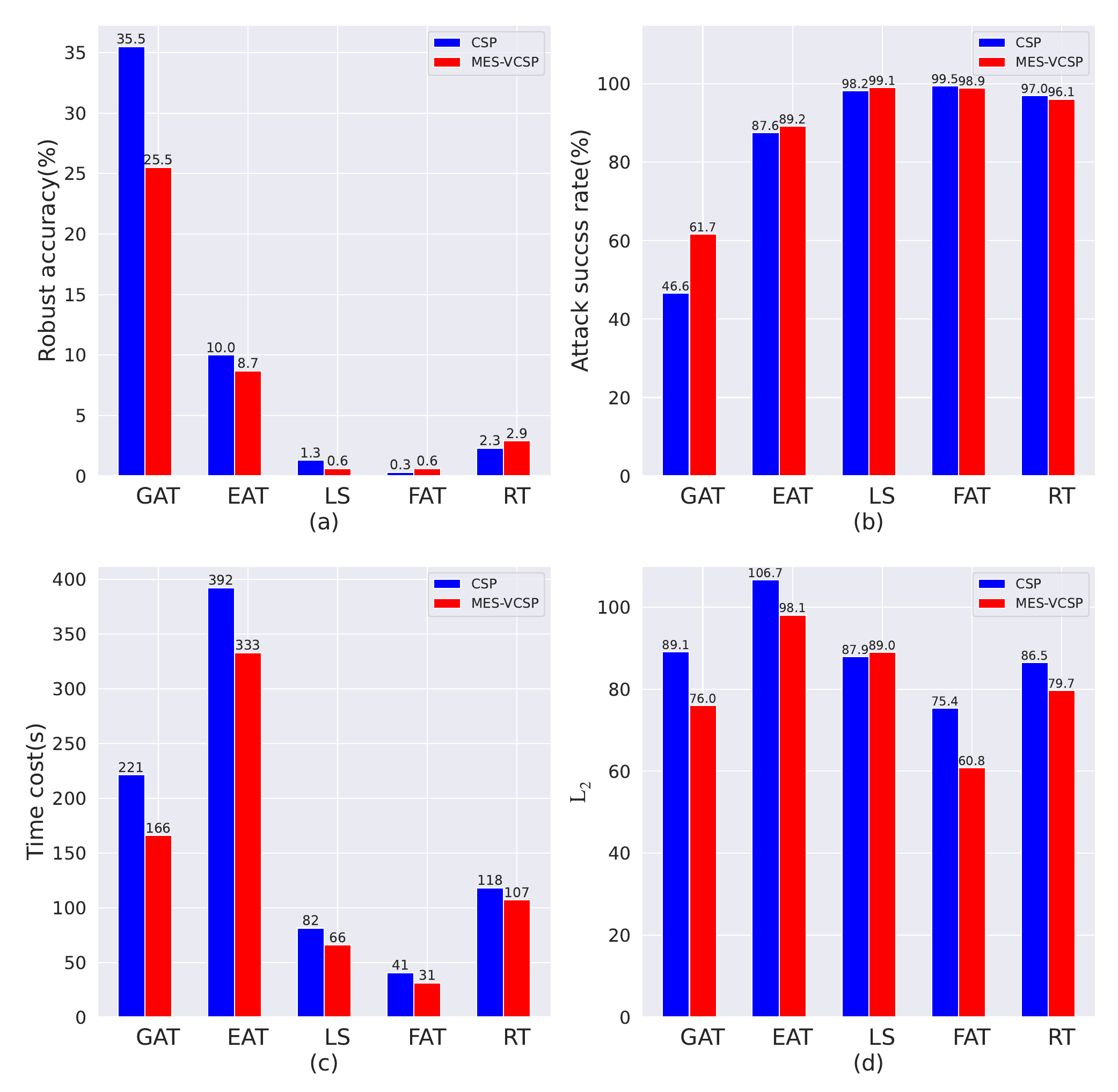}
\caption{The comparison of CSP and MES-VCSP on defensed model on the ImageNet dataset under the scheduled mode.}

\label{defenseimagenetsche}
\end{figure*}
During the process of evaluating the robustness of defensed models, we directly transfer the searched attacks to standard training models to assess them. The comprehensive evaluation results under different modes and datasets are shown in Figure. \ref{defensecifar10}, Figure. \ref{defensecifar10sche}, Figure. \ref{defenseimagenet} and Figure. \ref{defenseimagenetsche}, respectively. From Fig \ref{defensecifar10} and Figure. \ref{defensecifar10sche}, we can see that GAT can greatly improve the robustness of DNN models against CSP and MES-VCSP, whose robust accuracy on the CIFAR10 dataset can reach over 50$\%$. We also can find that under the defensed strategies, our proposed MES-VCSP can still possess superior attack performance than CSP. The improvement of ASR on five models on the CIFAR10 dataset can reach 5$\%$ to 10$\%$. In addition, MES-VCSP takes less time cost and possesses more natural AEs. The results of Figure. \ref{defenseimagenet} and Figure. \ref{defenseimagenetsche} show that GAT is the most effective defensed strategy. Under different defense strategies, MES-VCSP can still possess higher ASR than CSP. Among them, the improvement of MES-VCSP over CSP is the greatest on the GAT technique, which is about 15$\%$ to 20$\%$. As for the FAT technique, MES-VCSP is slightly better than CSP. These experimental results on different defensed strategies show that our proposed MES-VCSP can possess better attack generality than CSP.

\subsection{Analysis about the reason why VCSP is effective}

The effectiveness of MES-VCSP motivates us to conduct experiments to explore the reason for improving ASR greatly. We first select one defensed model on the CIFAR10 dataset as the threat model to conduct the rotation attack. The reason for selecting a rotation attack is that the search space, namely the lower and upper bound of the original variable, does not change if multiple attacks are performed using our proposed variable-length strategy. [2] stands for the rotation attack, where the interval of the rotation angle is [-10$^{\circ}$, 10$^{\circ}$]. [2, 2] means that we perform rotation attack twice, where the interval of the rotation angle is [-5$^{\circ}$, 5$^{\circ}$] at each time. The iteration number denotes the iteration number of each rotation attack. Table \ref{reson} presents the performance of different roration attacks. We can find that by increasing the length of the attack sequence, the attack performance becomes better, whose results correspond with the previous experimental results. The most important reason is that when we perform each semantic attack, we first randomly sample a value from the interval. Then based on the initial value, the gradient-based optimization is conducted to find the near-optimal perturbation value. Therefore, by the proposed variable-length strategy, we can sample multiple times the initial values, which can prevent the gradient-based optimization algorithm from being trapped in the local optimum. Another reason is that when we perform one type of semantic into multiple attacks, the interval has changed compared with the original one. Taking the Hue attack as an example, the perturbation interval is [-$\pi$, $\pi$] if it is performed only once time. However, if it is performed twice times, the second perturbation interval [-$\pi$/2, $\pi$/2] is on the basis of the perturbed image other than the original image. Hence, the total perturbation does not belong to [-$\pi$/2, $\pi$/2] strictly. But in any case, the generated AEs are more natural with smaller $L_{2}$ distance, which can illustrate the effectiveness of the proposed VCSP model.

\begin{table}[htbp]
\renewcommand\arraystretch{1.5}
\scriptsize
\centering
\caption{The performance of different rotation attacks.}
\label{reson}
\setlength{\textwidth}{24mm}{
	\begin{tabular}{ccccccc}
		
		\toprule
		
		\multicolumn{1}{c}{Iteration number} &\multicolumn{1}{c}{Rotation} &Robust Accuracy &\multicolumn{1}{c}{ASR}&Time cost &\multicolumn{1}{c}{$L_{2}$}\\
		
		\midrule			
		2&	[2]&61.3&	8.3& 48.4  & 58.7\\
		1&[2]&61.6&	8.3 & 35.9  & 65.5 \\
		1&	[2,2]&	61.0& 8.7 &  35.6&49.9\\	
		1&[2,2,2]&	\textbf{60.3}&\textbf{9.7} &  \textbf{37.4}&\textbf{49.0}\\	
		
		\bottomrule		
\end{tabular}}
\end{table}

%\subsection{The performance of MES-VCSP on different search spaces}
%
%We also investigate the transfer performance of NS-VCSP across different datasets. The searched attack on CIFAR10 dataset is evaluated on the ImageNet dataset. The ResNet50 model trained by Fast-AT on ImageNet dataset is taken as an example, the ASR on different search spaces is listed in Table \ref{transfer}. 
%
%\begin{table}[H]
%	\renewcommand\arraystretch{1.5}
%	\scriptsize
%	\centering
%	\caption{The transfer performance of the searched attack}
%	\label{transfer}
%	\setlength{\textwidth}{24mm}{
%		\begin{tabular}{ccccc}
%			
%			\toprule
%			
%			\multicolumn{1}{c}{Mode} &\multicolumn{1}{c}{[0,1,2]} &\multicolumn{1}{c}{[0,1,2,3]}&\multicolumn{1}{c}{[0,1,2,3,4]}\\
%			
%			\midrule			
%			fixed&	23.2 & \textbf{38.4}  & \textbf{95.1} \\
%			scheduled&	[0,1,2]& scheduled &  50.6\\		
%			
%			\bottomrule		
%	\end{tabular}}de
%\end{table}

\subsection{The performance of neighborhood search}
To illustrate the effectiveness of neighborhood search in the proposed MES-VCSP, we conduct the ablation study. We utilize random search to find the optimal attack sequence in CSP and VCSP, which are denoted as RS-CSP and RS-VCSP, respectively. Neighborhood search for CSP is denoted as NS-CSP. We select the Fast-AT model on the ImageNet dataset as the threat model. The robust accuracy and ASR of different methods are listed in Table \ref{abation}. From Table \ref{abation}, we can see that the proposed NS-VCSP can obtain the best performance. Neighborhood search can obtain better solutions than random search in both CSP and VCSP, showing the effectiveness of the optimization algorithm. By the same optimization algorithm, VCSP achieves higher ASR than CSP, which illustrates the effectiveness of our proposed variable-length composite semantic perturbations.

\begin{table}[H]
\renewcommand\arraystretch{1.5}
\scriptsize
\centering
\caption{The performance of the ablation study.}
\label{abation}
\setlength{\textwidth}{24mm}{
	\begin{tabular}{ccccc}
		
		\toprule
		
		\multicolumn{1}{c}{Method} &\multicolumn{1}{c}{Attack sequence} &Robust Accuracy ($\downarrow$)&\multicolumn{1}{c}{ASR ($\uparrow$)}\\
		
		\midrule			
		RS-CSP&[2,0,1,3,4]&	50.8 & 25.8  & \\
		NS-CSP&[3,4,2,0,1]&	46.5 & 32.1  &  \\
		RS-VCSP&	[3,4,2,0,1,3,2,1]& 38.6 &  43.6\\	
		NS-VCSP&	[0,1,2,3,4,2,3,1]& \textbf{30.8} &  \textbf{55.0}\\	
		
		\bottomrule		
\end{tabular}}
\end{table}

\subsection{The performance of hyperparameters}
\begin{figure}[htbp]
\centering
\includegraphics[scale=0.5]{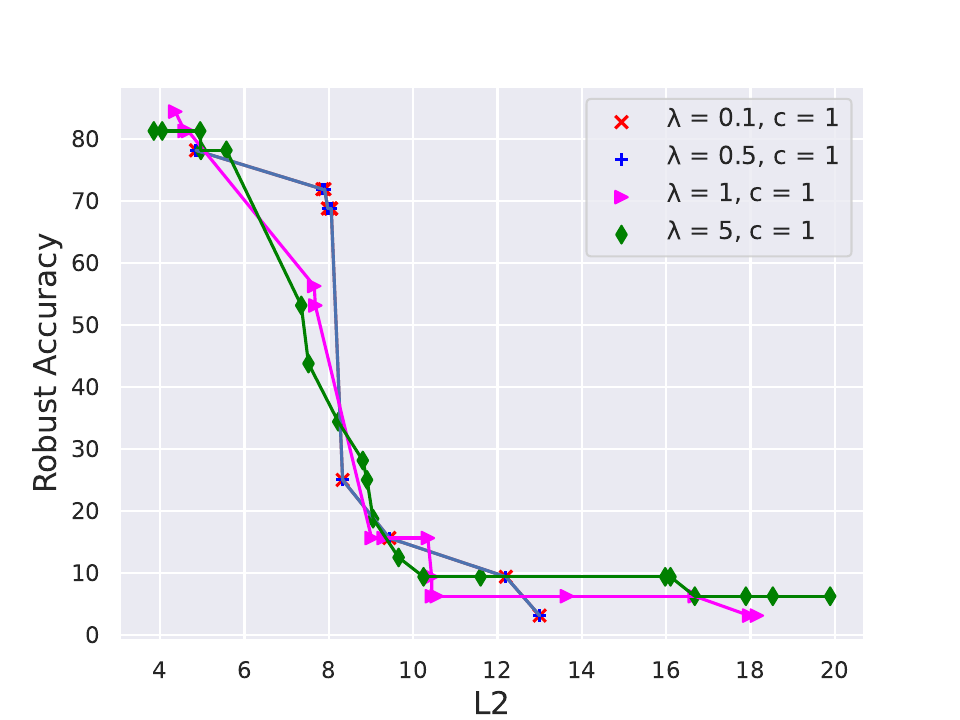}	
\includegraphics[scale=0.5]{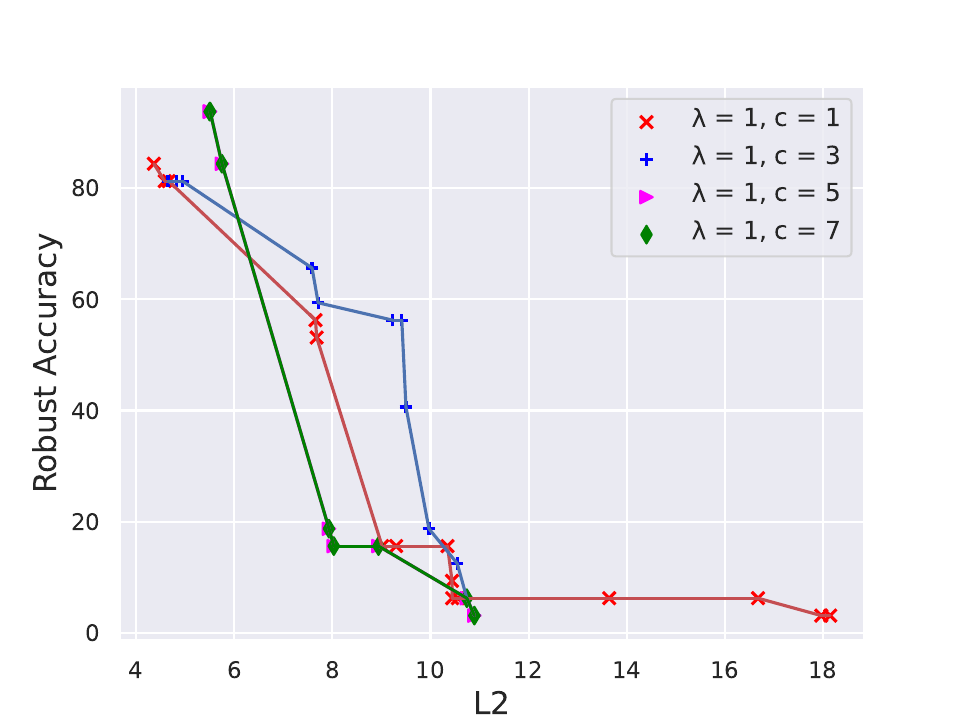}

\caption{The searched results under different hyperparameters.}

\label{hyper}
\end{figure}

During the optimization, the neighborhood search includes two hyperparameters, namely $\lambda$ and $c$. Hence, in this section, we investigate the performance of the whole algorithm under different hyperparameters. We set $\lambda$ to four values, including 0.1, 0.5, 1, and 5, respectively. We also set $c$ to 1, 3, 5, and 7. Other parameters, such as population size, are set the same as in the above experiments. The searched Pareto fronts are presented in Figure. \ref{hyper}. When $c=1$, we find the searched results of $\lambda=0.1$ and $\lambda=0.5$ are the same. Smaller $\lambda$ makes the searched results approach the robust accuracy, and the larger one would obtain the Pareto front that approaches $L_{2}$ distance. A similar phenomenon can also be seen in the experimental results under different $c$ values. When $\lambda=1$, we find the searched results of $c=5$ and $c=7$ are the same. Larger $c$ can help to find the attack sequence with lower robust accuracy and smaller $L_{2}$ distance, which means neighborhood search takes more iterations. In general, to obtain the satisfied results, both $\lambda$ and $c$ are set to 1 is a competitive choice.

\subsection{Discussions}
Different from previous works that only focus on studying improving the performance of semantic perturbations by manual design, our experimental results have revealed that the proposed MES-VCSP can automatically find the near-optimal variable-length composite semantic perturbations. In general, our method has two strengths. On the one hand, MES-VCSP reduces the human burden of designing the semantic adversarial attack method. On the other hand, the proposed method can obtain the AEs with a higher ASR, less time cost, and more naturalness. Experimental results have verified the effectiveness of the proposed mathematical model of variable-length CSP and search strategy. However, the proposed method still has some limitations. The defined search space is relatively simple. We believe designing a more diverse and efficient search space is one key component to influence the performance of automatically searching for composite semantic perturbations.

\section{Conclusion}
\label{conc}

This paper proposes a multi-objective evolutionary search of variable-length composite semantic perturbations. Different from the existing composite semantic perturbations method with the fixed length, MES-VCSP tries to search for a near-optimal variable-length attack sequence. Specifically, variable-length CSP constructs the mathematical model by adaptively adjusting the interval with the changing of the attack sequence, where one type of attack is performed multiple times. In addition, the multi-objective optimization algorithm consisting of NSGA-II and neighborhood search algorithm is introduced to find the optimal sequence. Experimental results on different models, including standard training models and defensed models, show that the searched attack by MES-VCSP can possess a higher attack success rate than CSP with less time cost and more naturalness.

The experimental results show that our proposed variable-length composite semantic perturbations and devised search strategy can obtain superior attack performance on various models. But how to design a more efficient and diverse search space towards further improving the attack performance against DNN models with more defensed techniques can be further studied.
\section*{Acknowledgments}
This work was supported by the National Natural Science Foundation of China (Nos.52005505).

\bibliography{mybibfile}

\end{document}